\documentclass[final,5p,times,twocolumn,authoryear]{elsarticle}

\usepackage{caption}
\usepackage[hidelinks]{hyperref}

\begin{document}

\journal{SMCDC23}
\renewcommand*{\today}{July 28, 2023}

\begin{frontmatter}

\title{
High-Throughput Phenotyping using Computer Vision and Machine Learning \\
\large SMC Data Challenge \#1
}

\author[fhs]{Vivaan Singhvi}
\author[fhs]{Langalibalele Lunga}
\author[fhs]{Pragya Nidhi}
\author[fhs]{Chris Keum}
\author[fhs]{Varrun Prakash}

\affiliation[fhs]{organization={Farragut High School},
            addressline={11237 Kingston Pike},
            city={Knoxville},
            postcode={37922},
            state={Tennessee},
            country={United States}}

\begin{abstract}
High-throughput phenotyping refers to the non-destructive and efficient evaluation of plant phenotypes. In recent years, it has been coupled with machine learning in order to improve the process of phenotyping plants by increasing efficiency in handling large datasets and developing methods for the extraction of specific traits. Previous studies have developed methods to advance these challenges through the application of deep neural networks in tandem with automated cameras; however, the datasets being studied often excluded physical labels. In this study, we used a dataset provided by Oak Ridge National Laboratory with 1,672 images of Populus Trichocarpa with white labels displaying treatment (control or drought), block, row, position, and genotype. Optical character recognition (OCR) was used to read these labels on the plants, image segmentation techniques in conjunction with machine learning algorithms were used for morphological classifications, machine learning models were used to predict treatment based on those classifications, and analyzed encoded EXIF tags were used for the purpose of finding leaf size and correlations between phenotypes. We found that our OCR model had an accuracy of 94.31\% for non-null text extractions, allowing for the information to be accurately placed in a spreadsheet. Our classification models identified leaf shape, color, and level of brown splotches with an average accuracy of 62.82\%, and plant treatment with an accuracy of 60.08\%. Finally, we identified a few crucial pieces of information absent from the EXIF tags that prevented the assessment of the leaf size. There was also missing information that prevented the assessment of correlations between phenotypes and conditions. However, future studies could improve upon this to allow for the assessment of these features. The use of machine learning and computer vision in high-throughput phenotyping has shown to be effective in analyzing large plant datasets, leading to a more comprehensive phenotype analysis in plants and showing potential in various agricultural and environmental applications.
\end{abstract}

\begin{keyword}
computer vision \sep machine learning \sep high-throughput phenotyping
\end{keyword}

\end{frontmatter}

\section{Introduction}
\label{intro}

\subsection{Background Information}
\label{background}

High-throughput phenotyping is defined as a breakthrough technology used in plant biology and agriculture to examine and assess plants' "anatomical, ontological, physiological, and biochemical features" through the use of images. Previous studies have shown its potential as a noninvasive replacement for traditional on-field techniques used to extract important phenotypic data. In recent years, this potential has been further intensified by coupling image-based phenotyping techniques with machine learning algorithms. This new approach has enabled the extraction of phenotypes from "complex" plant image datasets that were previously challenging to analyze efficiently. Furthermore, newer studies have shown that the technology has potential to identifying correlations between phenotype, genotype, and environmental metadata. 

\subsection{Research Objective}
\label{objective}

In this study, we aim to develop an accurate machine learning model for high-throughput phenotyping of leaves’ morphological traits (e.g. leaf size, shape, color) in plant images; unlike datasets in previous research, the dataset used in this study contains physical labels within the images, each containing important information on treatment (control or drought), block, row, position, and genotype. As a result, this study will address a new aspect of image-based phenotyping, namely, extracting data from the white labels to identify correlations between leaves’ phenotypes and data embedded within the white labels. During this project, we aim to answer the following questions: 

\begin{enumerate}

\item Is it possible to use optical character recognition (OCR) or machine learning techniques to “read” the label on each tag and generate a spreadsheet containing the treatment, block, row, position, and genotype? Doing this would dramatically simplify data collection, as this information is usually collected manually.

\item Can machine learning differentiate and classify different leaf morphologies among genotypes by classifying leaf shape or color characteristics?

\item Can a predictive model be built using leaf morphology classifications that may indicate that a particular genotype was cultivated in a “drought” or “control” condition?

\item GPS and other camera information are encoded in EXIF tags. Can this data be used to determine characteristics such as leaf size? Can other data, such as soil maps, weather, etc. be used to find correlations among phenotypes?

\end{enumerate}

\section{Related Works}
\label{related_works}

Traditionally, plant morphologies would be analyzed manually, wasting time and potentially damaging the environment. However, current breakthroughs in high-throughput phenotyping led by machine learning techniques \citep{koh2021automated, pound2017deep} and computer vision with deep learning \citep{mochida2019computer} have allowed for efficient and timely analyses of plant phenotypes. With these methods, researchers can extract features from plants and classify their morphologies effectively while having little effect on the plant itself.

\subsection{Advancements in High Throughput Phenotyping Techniques}
\label{related_works_1}

Advancements in the field of plant phenotyping have been pioneered by technology. From using hardware, such as Raspberry Pi imaging systems \citep{tausen2020greenotyper} to thermographic sensing techniques \citep{walter2015plant}, research in the field of high-throughput phenotyping has led to an acceleration in working efficiency. Plants are now able to be analyzed in fast, effective, and accurate methods allowing for them to be used in scientific discoveries. In our study, we rely on machine learning techniques to classify morphologies based on a plant image dataset provided.

\subsection{Machine Learning Approaches for Phenotypic Analysis}
\label{related_works_2}

Machine learning has enabled researchers with computer vision, allowing them to create programs that “read” image information in real-time. In addition, it increases the efficiency of obtaining leaf morphology data. However, researchers have found that traditional classification techniques, such as Random Forest Classification, have high performance but do not generalize well across datasets \citep{pound2017deep}. In order to increase the accuracy of models and the generalization of the model, deep learning techniques have been introduced. With deep learning models, models train themselves iteratively until they reach a desired outcome. These techniques, such as convolutional neural networks \citep{koh2021automated, pound2017deep} have shown to be highly accurate in classifying plant structures while maintaining performance across different datasets. Convolutional neural networks consist of several layers, which allow for a more discriminative and detailed analysis of plant images, leading to highly accurate classifications. In our study, we use traditional methods to classify plant morphology, as our focus is performing classifications with a single dataset rather than creating models highly transferable across datasets.

\subsection{Applications of High Throughput Phenotyping with Machine Learning}
\label{related_works_3}

Due to global events, such as climate change and global population increase, the ability to produce large amounts of healthy crops will be crucial to society. With machine learning being used for phenotyping, researchers will be able to rapidly analyze food crops to maximize production and breeding \citep{arya2022deep, shakoor2017high}.

As climate change and the global population increase, global crop yield will have to increase to provide for the growing populations. Further research has shown that climate change will continue to negatively affect crops and cause crop diseases \citep{newton2011implications}, hence the need for effective and precise analysis of plants, which can be done with machine learning models. In our study, we aim to show how machine learning can be used to organize plant data, dissect plant images into meaningful classifications, and allow for rapid investigation of plants for scientists in the field of plant phenotyping.

\section{Methodology}
\label{method}

Given the nature of the project, there are many aspects requiring obtaining large files, such as the dataset itself or large models. Therefore, all scripts needed to fully set up the project (downloading large files or setting up a large virtual environment), are viewable in the \verb|scripts/| folder of the GitHub repository, linked in Section \ref{code}.

All solutions to the challenge were implemented in Python 3.9/3.10. Several notable libraries used throughout the project include: OpenCV \citep{opencv_library}, the open-source computer vision library; Scikit-Learn \citep{scikit-learn}, Python implementations of dozens of machine learning algorithms and data processing tools; Pandas \citep{mckinney2010data}, a data manipulation library featuring the powerful \verb|DataFrame| object; and NumPy \citep{harris2020array}, an array manipulation library crucial for fast operations on images.

All models used in this project are trained and tested on data split by Scikit-Learn's \verb|train_test_split|. Additionally, Scikit-Learn's \verb|accuracy_score| is used to determine the accuracy of said models.

\subsection{Reading Labels with Optical Character Recognition}
\label{step_1}

To read the text, a pre-trained optical character recognition (OCR) model seemed like the optimal choice. However, there are several high-performing models available for use. The three candidates chosen for this project were \verb|PyTesseract| \citep{lee2017pytesseract}, \verb|EasyOCR| \citep{jaided2020easyocr}, and \verb|PaddleOCR| \citep{du2020ppocr}.

The three models were tested using metrics of performance and efficiency, which were measured through their accuracy score and time taken on a sample dataset of nine images. The results of the testing can be viewed in \ref{app_ocr_testing}.

\verb|PaddleOCR| was superior in speed, being significantly faster than \verb|EasyOCR| and marginally faster than \verb|PyTesseract|. It is also more accurate in both measurements, having a higher accuracy score with and without null values.

In order to fix the aforementioned null values from the OCRs, images went through augmentation, with subsequent attempts being made to read text during each step. First, images are rotated 45 degrees in order to fix potential orientation issues. Then, the original image is thresholded using OpenCV's \verb|adaptiveThreshold| to amplify the edges. Finally, if nothing works, the image is both rotated and thresholded.

\begin{center}
    \includegraphics[width=0.45\textwidth]{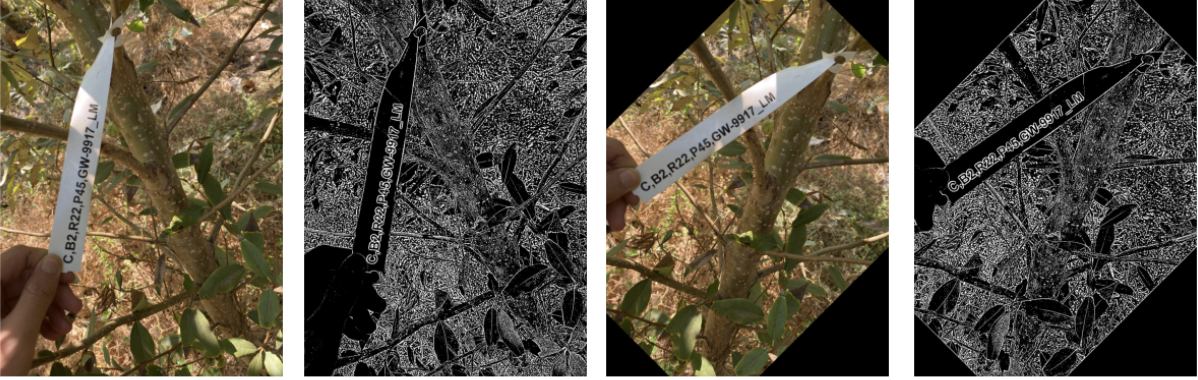}
    \captionof{figure}{Different versions of images tried for optical character recognition, in order}
\end{center}

After the text was read by the OCR, regular expressions (RegEx) were used to find the precise details mentioned in the dataset: treatment, block, row, position, and genotype. A subjective analysis of the images in the dataset showed that the treatment was either a \verb|C| or a \verb|D|, and the remaining features followed the RegEx patterns below (note: \verb|\d| represents a digit, anything between 0-9) :

\begin{verbatim}
  Block:      B\d+
  Row:        R\d+
  Position:   P\d+
  Genotype:   [A-Z]{2,}(-\d+)+(_\d+)*(_[A-Z]+)?
\end{verbatim}

After all the text is read and processed, it is converted to a Pandas \verb|DataFrame| and saved to an Excel spreadsheet, which will be used in the following steps. Then, 30 random images are selected and the accuracy of the results is tested by comparing inputted to read features. As with the OCR testing earlier, it is calculated both with and without unread values.

\subsection{Classifying Leaf Morphologies with Image Processing}
\label{step_2}

In order to be able to determine the morphological characteristics of the plants, it was necessary to perfectly segment the leaves from each image. This proved to be a challenging endeavor due to the complexity of the background of the images; other datasets, such as the Komatsuna dataset, \citep{uchiyama2017easy} had 'cleaner' image backgrounds.

To obtain perfect masks for leaves in each image, the Segment Anything Model (SAM) \citep{kirillov2023segment} was the obvious choice, due to its groundbreaking accuracy and ease of use. Being a relatively new model, this paper is one of the first to utilize a pre-trained model of its caliber.

The \verb|AutomaticMaskGenerator| provided generates accurate masks for everything in an image. However, it has two issues: it may generate too many masks (especially for images as cluttered as those in the dataset), and is computationally expensive, taking too long to be a plausible approach for all 1672 images in this dataset. Through a \verb|SamPredictor| in conjunction with an ONNX \citep{bai2019onnx} \verb|InferenceSession|, masks would generate much quicker, and only from desired points. However, to obtain these desired points, the images would have to be processed to approximate the locations of the leaves.

The first step was to hide non-leaf elements in the image. A popular technique called HSV filtering was employed, used in a similar project by \cite{szachowicz2021komatsuna}. This involves filtering all pixels that are not between an HSV-encoded color range, helping eliminate many background objects such as wood, dirt, and of course, the label.

Then, OpenCV's implementation of the \verb|Canny| algorithm was used to highlight significant edges in the image, which were present around the boundary of leaves but were minimal inside of them. However, some leaves had boundary edges with gaps, causing most contours detected (using OpenCV's \verb|findContours|) to be sporadic. Therefore, the edges were dilated in multiple iterations by using a large kernel, causing the boundaries to be closed. After contours were found on the resulting image, they were filtered by pruning those that had a height or width (found using the \verb|boundingRect| function around each contour) too small compared to that of the image. Additionally, contours with not enough green inside them (namely less than $\frac{100}{255}$ green) were removed.

\begin{center}
	\centering
	\includegraphics[width=0.23\textwidth]{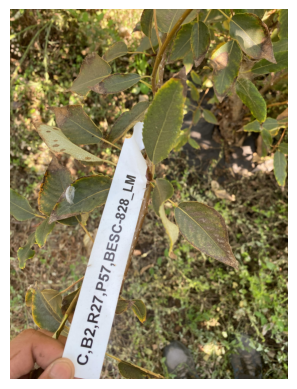}	
        \includegraphics[width=0.23\textwidth]{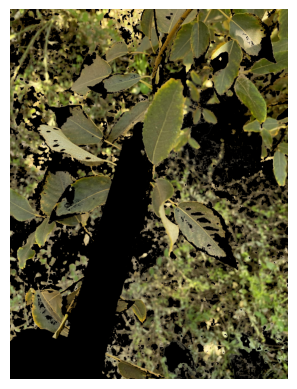}      	\includegraphics[width=0.23\textwidth]{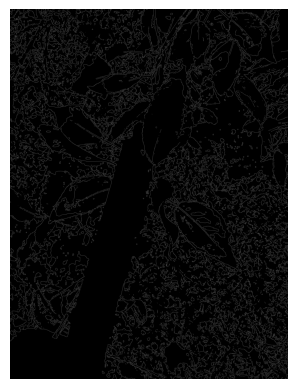}	
	\includegraphics[width=0.23\textwidth]{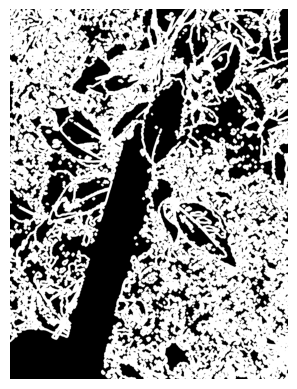}	
	\includegraphics[width=0.23\textwidth]{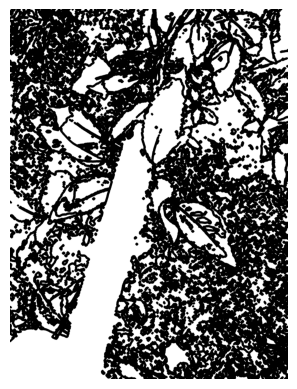}	
	\includegraphics[width=0.23\textwidth]{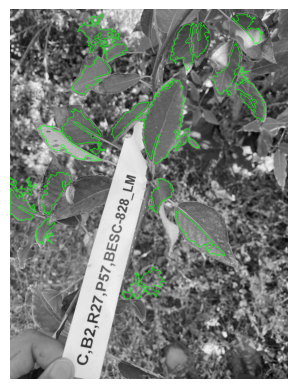}	
	\captionof{figure}{Pipeline of image transformations for leaf approximation, read left to right. The last image visualizes detected contours}
\end{center}

To find the approximate points of these leaf contours, the midpoint was calculated using this simple formula, with $x, y, w, h$ being the outputs of OpenCV's \verb|boundingRect| function:

\begin{equation}
    \mbox{midpoint} = (x + \frac{w}{2}, y + \frac{h}{2})
\end{equation}

By using these points as target points for the SAM/ONNX mask predictor, leaves could reliably be obtained from the image. Masks could then be saved for each image by converting each of an image's masks to a random uniform grayscale value and combining them all into an image. While very accurate, however, some leaves detected by the program were not appropriate for use in morphology analysis. For example, some leaves could be cut off by another object, or, in rare cases, be green wood misinterpreted as leaves.

Thus, it was necessary to train a machine learning model that was able to accurately detect these kinds of 'bad' leaves. Training data was generated by showing randomly selected leaves and getting human input regarding their suitability, extracting their features, and storing the data in a file. The model was implemented using Scikit-Learn's \verb|RandomForestClassifier|, and had an accuracy of 90.91\% on testing data.

\begin{center}
	\centering
	\includegraphics[width=0.23\textwidth]{images/seg_pipeline/6.png}	
        \includegraphics[width=0.23\textwidth]{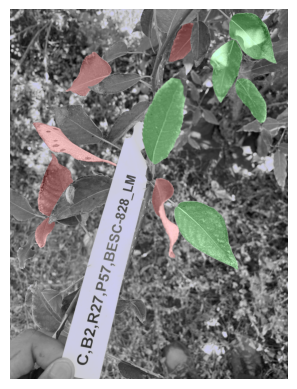}	
	\captionof{figure}{From leaf approximations with image processing to filtered leaf segmentations, from the SAM/ONNX Runtime (green leaves fit for classification)}
\end{center}

Then, each leaf could be cropped and rotated, which, when combined with masking every pixel outside its border, effectively highlighted and isolated it for visual purposes. By repeating this process for each image in the dataset, random leaves could then be selected to generate data for training the morphological classification model.

The features selected for morphology prediction were: color, between light-green, dark-green, yellow-green, and yellow; shape, between ovate, lanceolate, elliptical, and oblong \citep{nakano2020plant}; and level of brown splotches (representing the level of the withering of the leaf), between none, low, medium, and high.

\begin{center}
    \includegraphics[width=0.45\textwidth]{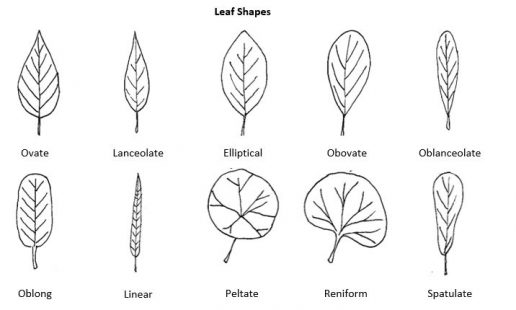}
    \captionof{figure}{Leaf shape reference from the article by \cite{nakano2020plant}}
\end{center}

By using a slightly modified version of the aforementioned feature extraction function in conjunction with 
a Scikit-Learn \verb|LabelEncoder| to prepare the target features, the classifier could be implemented with Scikit-Learn's \verb|MultiOutputClassifier|. This is a framework for using the same algorithm on three different features at once; namely, the algorithm used was \verb|XGBoost|, which seemed to trump other models in terms of accuracy.

After the model was trained, it was used on each image in the dataset by finding all eligible leaves and determining the mode classification after each eligible leaf was run through the aforementioned model. The results were saved in a copy of the spreadsheet from the previous step. 

\subsection{Treatment Predictions using Morphological Classifications}
\label{step_3}

The task for this step was to build a model to predict if a leaf was grown in a control or drought environment, given morphological classifications from the previous step.

The classifications were easily attainable because they were saved in a spreadsheet. However, the challenge for using classification data as features to train a model with is that they are categorical, rather than quantitative; data needed to undergo extra transformations to prepare it for use in predictions.

To properly represent the classifications, the method of one-hot encoding \citep{brownlee2020why} was implemented using the Pandas \verb|get_dummies| method. This is a process in which a feature is split into multiple columns, with each row having a one if it matches the column or a zero if it doesn't. 

\begin{center}
\resizebox{\columnwidth}{!}{%
\begin{tabular}{|c||c c c c|} 
 \hline
 leaf\_color & light\_green & dark\_green & yellow\_green & yellow  \\
 \hline
 light\_green & 1 & 0 & 0 & 0 \\
 yellow\_green & 0 & 0 & 1 & 0 \\
 dark\_green & 0 & 1 & 0 & 0 \\ 
 light\_green & 1 & 0 & 0 & 0 \\
 yellow & 0 & 0 & 0 & 1 \\
 \hline
\end{tabular}
}
\captionof{table}{Example transformation of a column with one-hot encoding, with the original column on the left and the encoded columns on the right}
\label{Table1}
\end{center}

This process was done with the leaf color and shape variables. However, since a leaf's level of brown splotches is ordinal, it was assigned levels 0, 1, 2, and 3, corresponding to 'none', 'low', 'medium,' and ‘high.' The data for expected treatments were obtained by reading rows of the spreadsheet that had treatments successfully read.

Now, using this data, a predictor could be trained and deployed. The chosen classifier for this task was Scikit-Learn's \verb|RandomForestClassifier|, as it has shown to be a capable model throughout the project. After the model was trained, it was deployed on the spreadsheet, predicting treatments for rows that lacked a properly read one.

\subsection{Using EXIF Data to Assess Leaf Size and Analyze Correlations Between Morphologies and Environments}
\label{step_4}

First, we aimed to measure the size of a leaf through metadata encoded in EXIF Tags, which are embedded within each of the images. These tags typically give important information about the images, including the type of camera utilized, the distance of the camera from a photo, and photo dimensions (in pixels). In order to initiate the extraction of such metadata from the images, the \verb|exif| package was used and gave a variety of retrieval options. 

Out of all the metadata extracted from the EXIF tags, the most notable pieces of information included \verb|ResolutionX|/\verb|ResolutionY| tags and \verb|FocalLength|. Although these pieces of information can assist in developing a mechanism to measure leaf size in the dataset, it simply does not suffice alone; additional information, such as how far the leaf was from the camera, was needed. With more information, such as \verb|FocalPlaneResolution| - showing the number of pixels per real unit (\verb|in| or \verb|cm|) - in addition to previously mentioned features, it is possible to assess the size of specific components of an image. For example, a study on maize phenotyping by \cite{liu2021pocketmaize} showed that embedded information in images was enough for a smartphone application to accurately determine measurements of plant height and leaf area in a photo.

Another approach we considered to measure leaf size included using the white labels, held up by the researcher in each image, as a known reference object. Although this method seemed promising, there were certain limitations in most images in the dataset that prevented a consistent measurement. First, the white label in the image was often obstructed by the researcher’s hands, as shown in \ref{app_blocked}. This obstruction would ultimately interfere with the alignment of the white label, making the label unreliable as an object of reference. Another limitation to this method came from the fact that the labels were often tilted, not in the same plane as the leaves. The tilt of the labels caused inaccuracies in measurement and would lead to misrepresentation of the true length. Along with that, there was no way of finding out how far forward or behind the leaf was relative to the tag, meaning the method would result in an inaccurate representation of the leaf. Again, this issue could have been resolved with more careful documentation of images without obstruction and a better-positioned label alignment.

In the second part of challenge 4, we aimed to find a correlation between phenotypes and environmental features, such as soil type and weather conditions. Using the EXIF tags, we were able to retrieve the coordinates where the image was taken. However, we noticed all of the images contained were taken in approximately the same location.

\begin{center}
    \includegraphics[width=0.45\textwidth]{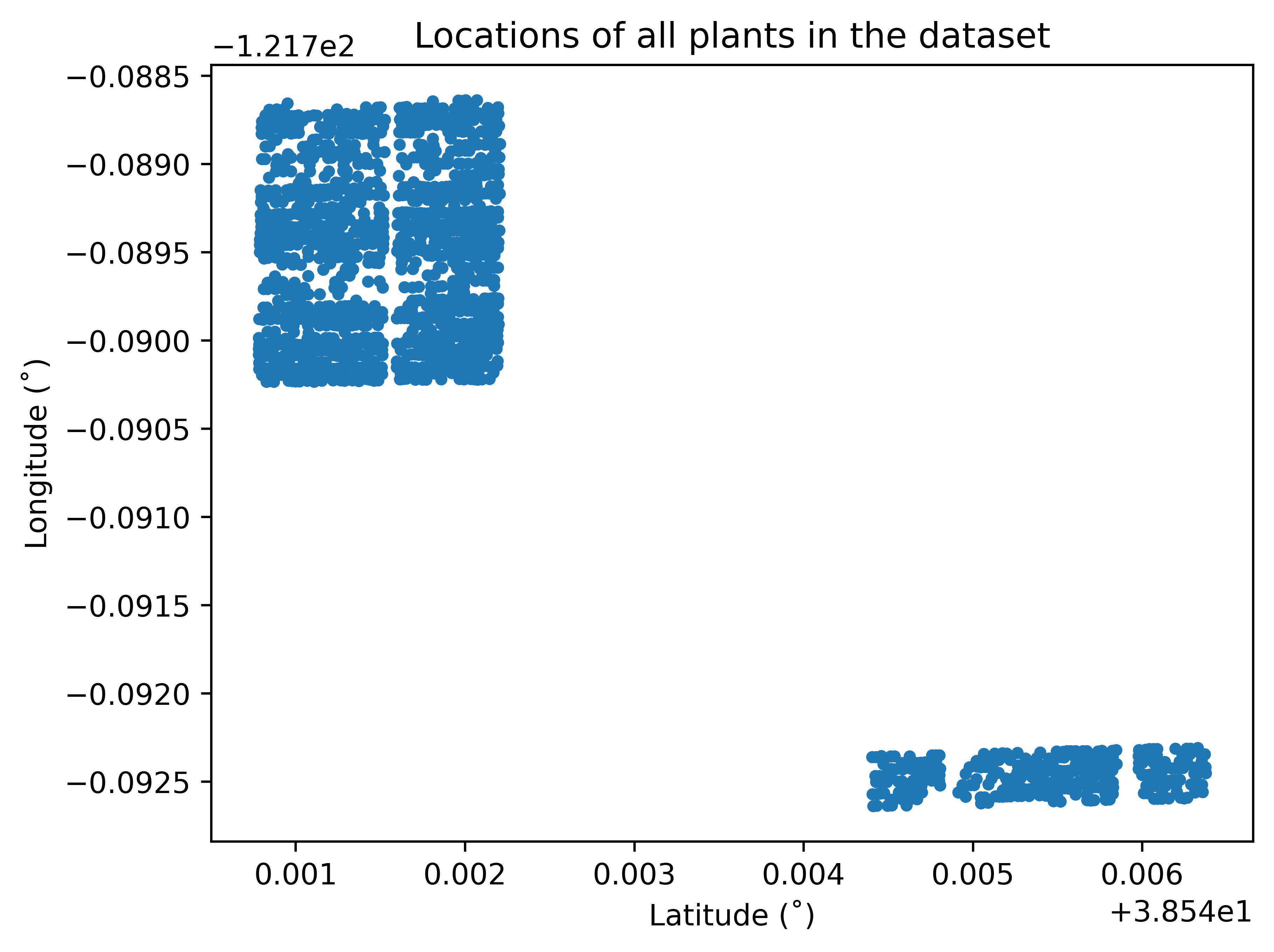}
    \captionof{figure}{Geographical plot of all plants using encoded EXIF GPS data.}
\end{center}

Therefore, any weather or soil API would be unable to provide specific conditions. If we were given this information as part of the dataset, we would have been better able to find correlations. Regardless, a study on how the temperature affects the phenotype and development plasticity in Arabidopsis Thaliana \citep{ibanez2017ambient} placed the plants in different temperature environments in order to measure their differences. If we had more information regarding the conditions in which the plants were raised, we would be better able to determine if there was a correlation between phenotypes and environment.

\section{Results and Discussion}
\label{results}

All confusion matrices used in this section originate from Scikit-Learn's \verb|confusion_matrix| and \verb|ConfusionMatrixDisplay|.

\subsection{Reading Labels with Optical Character Recognition}
\label{step_1}

The OCR was able to read 91.3\% of the labels in the images of the entire dataset. On the 30 test images, the model had a 77.33\% accuracy on extracted features (94.31\% with null values omitted). A subjective analysis showed that for all features except for genotype, the interpreted values seem to match the true values almost always.

\begin{center}
\resizebox{\columnwidth}{!}{%
\begin{tabular}{|c|c|c|c|c|c} 
 \hline
 filename & treatment & block & row & position & genotype \\
 \hline
    ... & D & 1 & 8  & 32 & BESC-34 \\
    ... & C & 1 & 10 & 12 & **BESC-417\_LM**,core \\
    ... & C & 2 & 3  & 40 & BESC-468 \\
    ... & C & 2 & 6  & 54 & BESC-28\_LM \\
    ... & C & 1 & 24 & 22 & **LILD-26-5\_LM**,core \\
 \hline
\end{tabular}
}
\captionof{table}{First five rows of data saved to the spreadsheet from step 1, with filenames omitted to conserve space}
\label{Table1}
\end{center}

Before the text was inserted into the above spreadsheet, it was present in a Pandas \verb|DataFrame|. Here, it was possible to analyze the statistics of the data using the \verb|info| method, containing useful information such as null value count:

\begin{verbatim}
    RangeIndex: 1672 entries, 0 to 1671
    Data columns (total 6 columns):
     #   Column     Non-Null Count  Dtype  
    ---  ------     --------------  -----  
     0   filename   1672 non-null   object 
     1   treatment  1098 non-null   object 
     2   block      1306 non-null   float64
     3   row        1388 non-null   float64
     4   position   1414 non-null   float64
     5   genotype   1431 non-null   object 
\end{verbatim}

Evidently, the former attributes were read the least compared to those near the end of the tag, with 66\% non-null values for the treatment and 86\% non-null values for the genotype. This is likely due to the many forms of obstructions in the images, such as a leaf, a data collector's hand, or even the boundary of the image itself. Examples of such can be seen in \ref{app_blocked}.

Even though the information suggests that there is more error in reading the treatment, it is more likely that it stems from text processing. The interpreted genotype has more room for error, as there is an extensive number of possibilities for it, while the treatment must be a \verb|C| or a \verb|D|, leaving little room for error.

Regardless, it is undeniable that the use of \verb|PaddleOCR|, a relatively new and underused model, proved significantly beneficial and successful in reading labels from the images in the dataset. However, one evident issue lies in the fact that the suffix of the genotype (such as \verb|_LM|) was often not read properly, causing the text processing to drop it entirely. Increased text pre-processing may have alleviated this issue.

\subsection{Classifying Leaf Morphologies with Image Processing}
\label{step_2}

The spreadsheet from the last step was successfully updated with morphological predictions from this step.

\begin{center}
\resizebox{\columnwidth}{!}{%
\begin{tabular}{c|c|c|c|c|} 
 \hline
 genotype & leaf\_color & leaf\_shape & brown\_splotches \\
 \hline
    BESC-34	&                  light\_green &   ovate &	    none     \\
    **BESC-417\_LM**,core &	   yellow\_green	&  ovate &      high     \\
    BESC-468 &	               yellow &        lanceolate &	high     \\
    BESC-28\_LM &	               dark\_green &	   ovate &      low      \\
    **LILD-26-5\_LM**,core &	   yellow\_green	&  lanceolate &	medium   \\
 \hline
\end{tabular}
}
\captionof{table}{A continuation of Table 2, with morphological data added from step 2}
\label{Table1}
\end{center}

The model had an average accuracy of 62.82\%, which, when considering the fact that each feature had four different classes, is a reasonable accuracy level. The confusion matrices generated by comparing test and predicted data for each feature are below:

\begin{center}
    \includegraphics[width=0.45\textwidth]{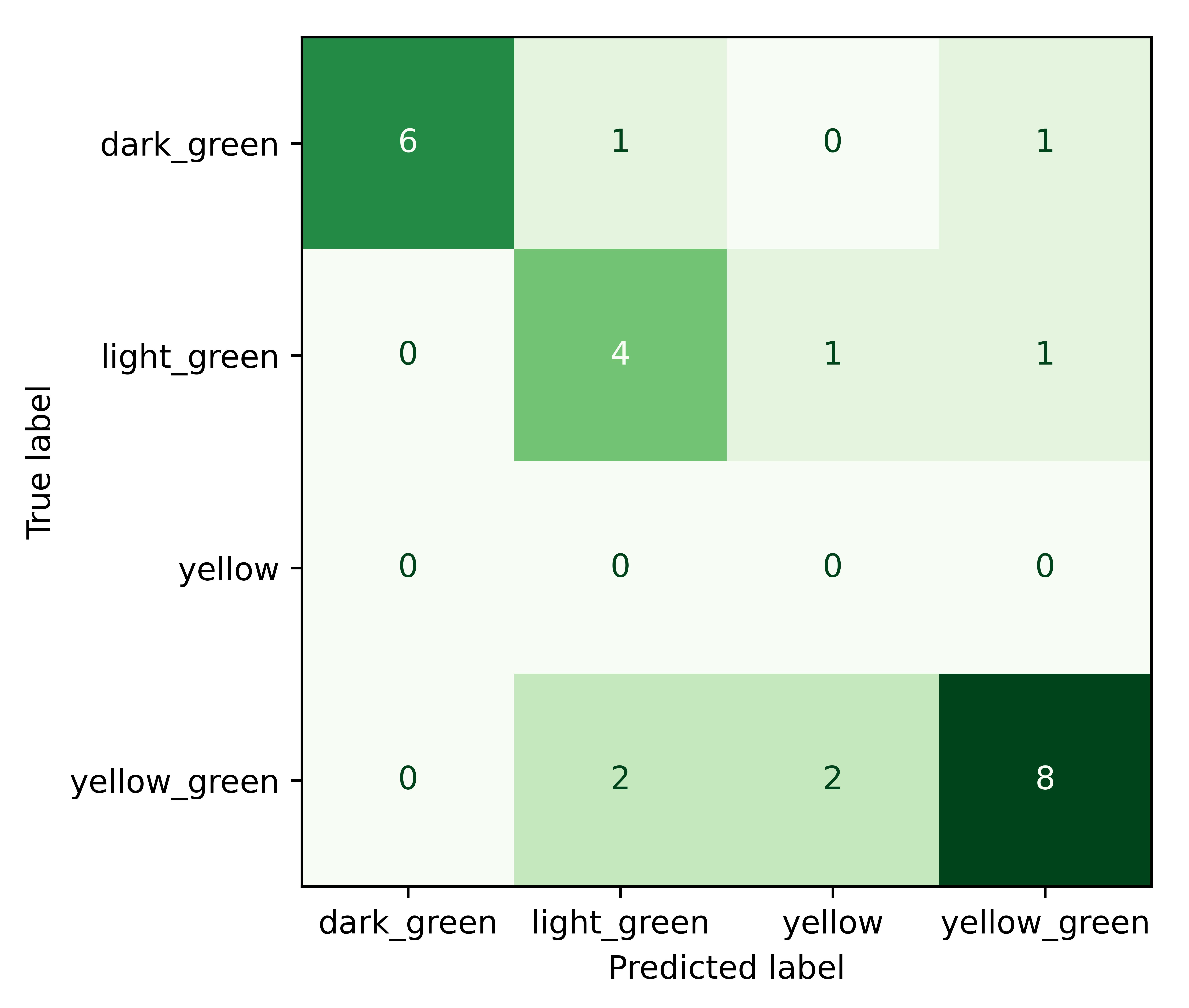}
    \captionof{figure}{Confusion matrix for testing data on the leaf color predictor}
\end{center}

This part of the model had an accuracy score of 69.23\%. It seems to be reliable in predicting leaf color but has difficulty classifying true yellow-green leaves; however, it still correctly predicts yellow-green most of the time. Because the testing data seems to lack yellow leaves, the model's accuracy regarding them is unknown.

\begin{center}
    \includegraphics[width=0.45\textwidth]{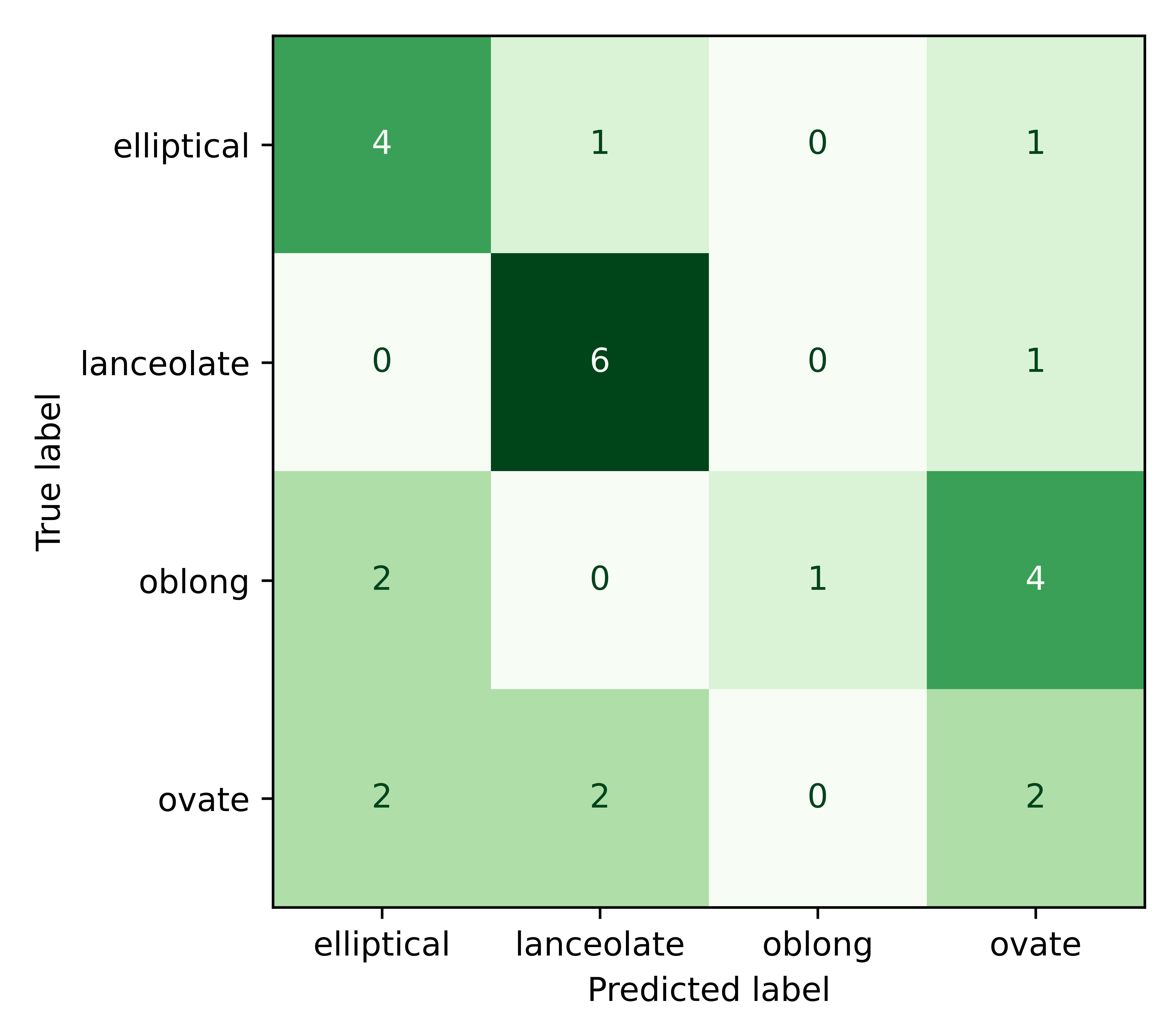}
    \captionof{figure}{Confusion matrix for testing data on the leaf shape predictor}
\end{center}

While mostly reliable with color, the model is sporadic in predicting leaf shape, with an accuracy of 50.00\%. When leaves are truly shaped as oblong or ovate, the predictions are essentially random, including many more wrong guesses than correct guesses. However, this may be partially due to the subtlety of the difference between leaf shapes, which may be difficult for a model to predict.

\begin{center}
    \includegraphics[width=0.45\textwidth]{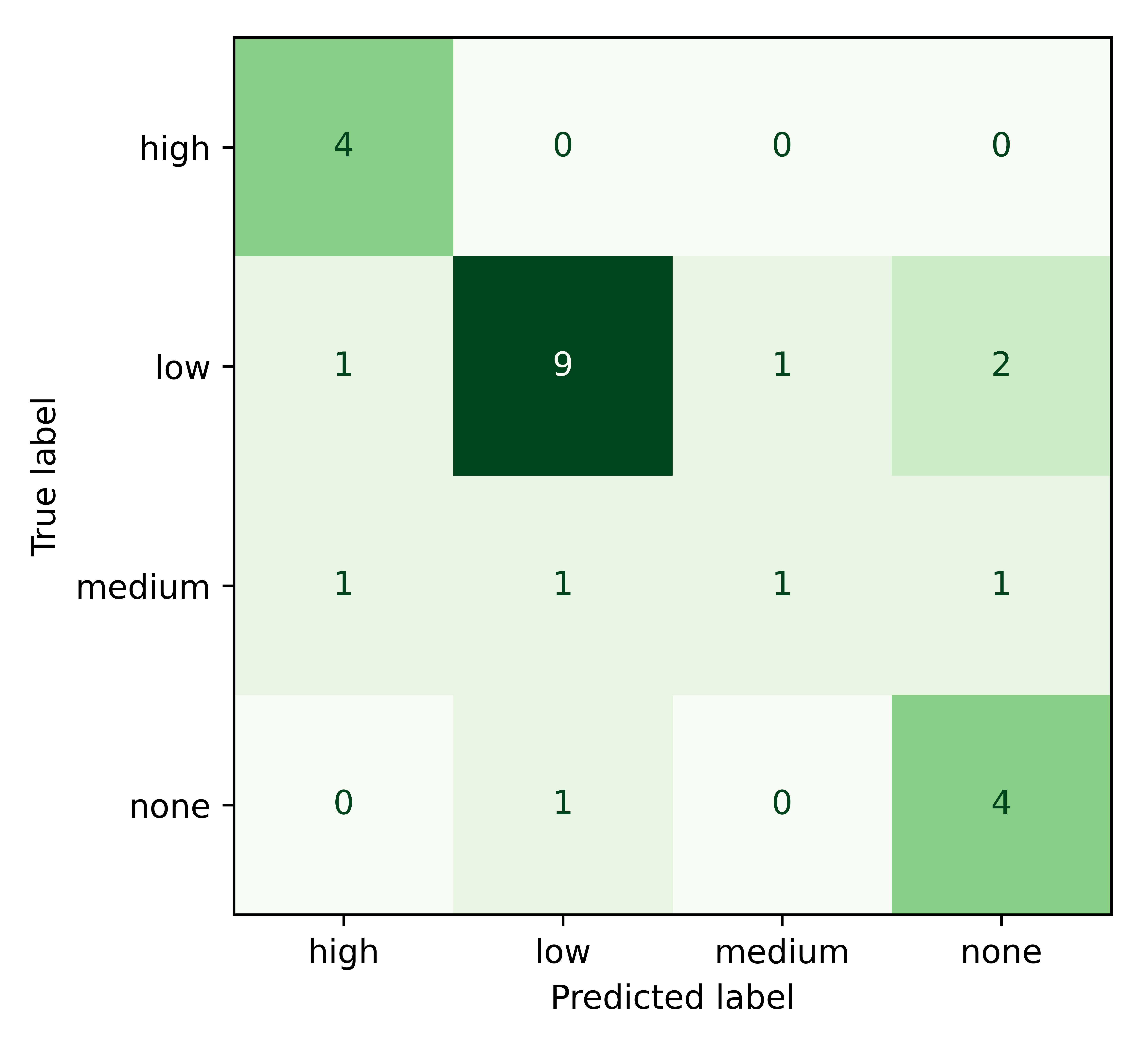}
    \captionof{figure}{Confusion matrix for testing data on the brown splotch level predictor }
\end{center}

While there is more 'noise' in the predictions for brown splotch levels when compared to those for leaf color, the model seems to have a moderate degree of accuracy (at 69.23\%), with the exception of classifying leaves with "medium" splotch levels. 

A source for much of the error in the model's predictions may lie in human error. For example, when labeling data, we may have incorrectly labeled any of the three features for a given leaf due to their aforementioned ambiguity. It is also worth noting that many of the classes were difficult to distinguish even subjectively, such as distinguishing between yellow-green and light-green colors or between elliptical and ovate leaf shapes. 

\subsection{Treatment Predictions using Morphological Classifications}
\label{step_3}

The predictive model on the treatment of the plants had an accuracy of 60.08\%. 

\begin{center}
    \includegraphics[width=0.45\textwidth]{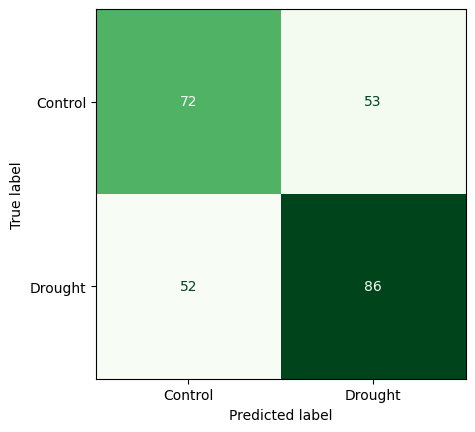}
    \captionof{figure}{Confusion matrix for the treatment prediction model}
\end{center}

While more predictions were true than false, it is undeniable that the model is unreliable for determinate results. The inaccuracy is likely due to several factors, including the fact that the only features used in the model were the morphological classifications from step 2, which may have been inaccurate to begin with. Additionally, there may have been overfitting on training data.

\subsection{Using EXIF Data to Assess Leaf Size and Analyze Correlations Between Morphologies and Environments}
\label{step_4}

Overall, the metadata embedded in the EXIF tags specific to the dataset did not prove to be enough for direct measurement of leaf size. Given more useful tags, such as \verb|FocalPlaneResolution|, the task may have been possible. Additionally, using the label as a reference for measurement also proved impossible due to the various augmentations of the labels in images. Future studies could include additional calibration features to enable the development of a more feasible method for leaf size assessment. For the second part of the challenge, we needed larger differences in locations or times that the images were taken in order to study how the conditions in those periods affected the phenotypes.

\section{Conclusions}

High throughput phenotyping has gained prominence due to its potential in solving a wide variety of agricultural problems. With the world population projected to reach 9.3 billion people by 2050 and a need to produce 60\% more food \citep{silva2012feeding}, it will become crucial for researchers to discover a method for identifying productive genotypes for plant breeding. With computer vision techniques, such as optical character recognition and image processing, and machine learning methods, such as classifiers to predict morphologies and treatments, productivity in the field of plant/crop production will increase rapidly, along with the accurate analysis of plants.

\subsection{Originality and Uniqueness}

This project is one of the first to utilize \verb|PaddleOCR| \citep{du2020ppocr}, an extremely lightweight and capable optical character recognition model, in the context of reading plant labels. Unlike other approaches, it is resilient to augmentations in data, such as label rotation or minor interruptions in labels.

It is also one of the first to use the Segment Anything Model (SAM) \citep{kirillov2023segment} in the context of leaf segmentation. When combined with image processing, leaves can now be extracted from an image with no training or processing steps for the user. The SAM allows for much more accurate masks to be obtained and is robust towards heavily noisy data such as ours. This is especially important when leaf shape needs to be classified, as accurate masks have more accurate features.

Through the use of pre-built classifiers like \verb|RandomForestClassifier|, this paper shows that such models can be useful despite the little effort used by the programmer.

\subsection{Significance and Specific Contributions}

By successfully applying an OCR to read labels, our research shows potential for automating a data extraction process, which can significantly reduce the manual labor needed to organize information. Large datasets can be processed within minutes, with minimal effort from the user.

The use of machine learning on leaves allows for easy analysis of plant morphologies, which reduces the need for subjective determinations. Although the model has room for improvement, the process shows promise and can be easily optimized. Moreover, using another model to predict the kind of environment a plant was grown, as was done in step 3, allows for even more valuable information to be obtained just from a single photograph. Researchers can save hours of effort analyzing thousands of images by utilizing tools such as these.

Even though the EXIF tags provided by the dataset proved insufficient for determining leaf size, with the proper additional information, this task could become trivial and greatly benefit researchers who would otherwise have to manually measure each leaf with a ruler.

\subsection{Limitations and Possible Improvements}

To achieve higher accuracy in OCR label reading, future research could use datasets that are more uniform with clearer labels. On a similar note, the images in the dataset could have a clear focus on a specific leaf, making leaf classification models better.

The dataset's lack of metadata prevented the use of EXIF tags to determine leaf size and correlations between phenotypes and environmental features. Therefore, a dataset with more information both embedded into images and embedded into separate files (such as files with soil condition information) could allow researchers to meet these objectives.

However, if the same dataset is used, future researchers could build more accurate and complex neural networks for classification. The models used in this paper were all prebuilt and not fine-tuned to our needs. Additionally, the image processing could have improved reliability, making data more fit for use in training or feature prediction.

\subsection{Future Directions}

Future research could be done to expand the species analyzed for morphology classifications. In this study, only Populus Trichocarpae were used, so the models used may only be effective on this species.

Since a pre-trained segmentation model (the SAM) was used in this study, researchers could attempt to build segmentation models fine-tuned to only recognize leaves, which could increase model efficiency and provide more consistent results.

\section{Code Availability}
\label{code}

All code is publicly available under the MIT License on GitHub here: \url{https://github.com/vivaansinghvi07/smoky-mountain-data-comp}.

\section*{Acknowledgements}
\label{acknowledgements}

We thank Dr. Ty Frazier at Oak Ridge National Laboratory for his helpful suggestions and mentoring throughout this project.

\bibliographystyle{elsarticle-harv}
\bibliography{main}

\clearpage
\appendix

\section{Results of OCR Model Testing}
\label{app_ocr_testing}

\begin{center}
	\includegraphics[width=0.45\textwidth]{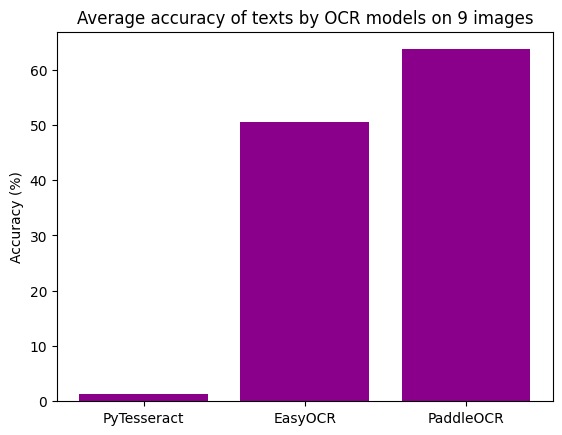}	
	\captionof{figure}{Average accuracy score for reading the sample dataset for the three OCR models}
\end{center}

\begin{center}
	\includegraphics[width=0.45\textwidth]{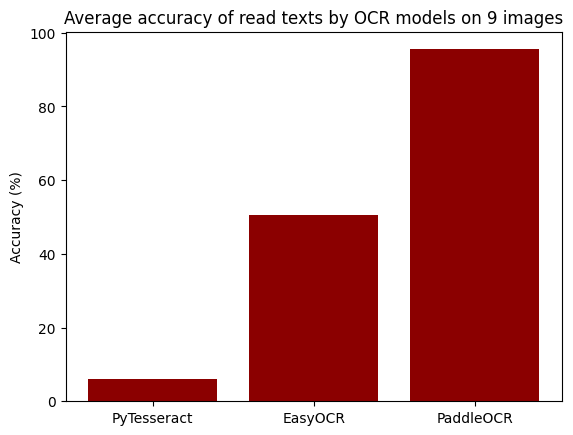}	
	\captionof{figure}{Average accuracy score only including non-zero values (only images that were able to be read)}
\end{center}

\begin{center}
	\includegraphics[width=0.45\textwidth]{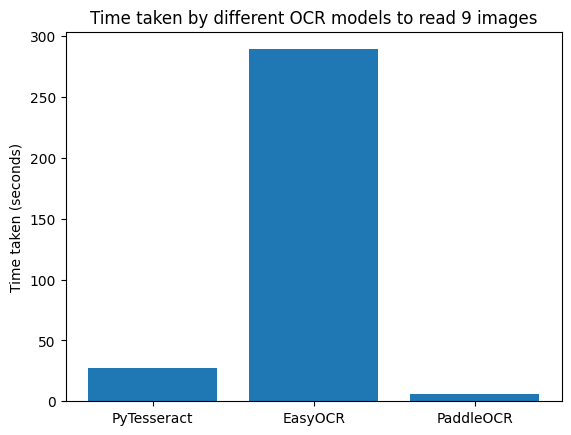}	
	\captionof{figure}{Total time taken for the three OCR models on the sample dataset}
\end{center}

\section{Examples in the Leaf Segmentation Process}
\label{app_leaf_seg}

\begin{center}
    \includegraphics[width=0.45\textwidth]{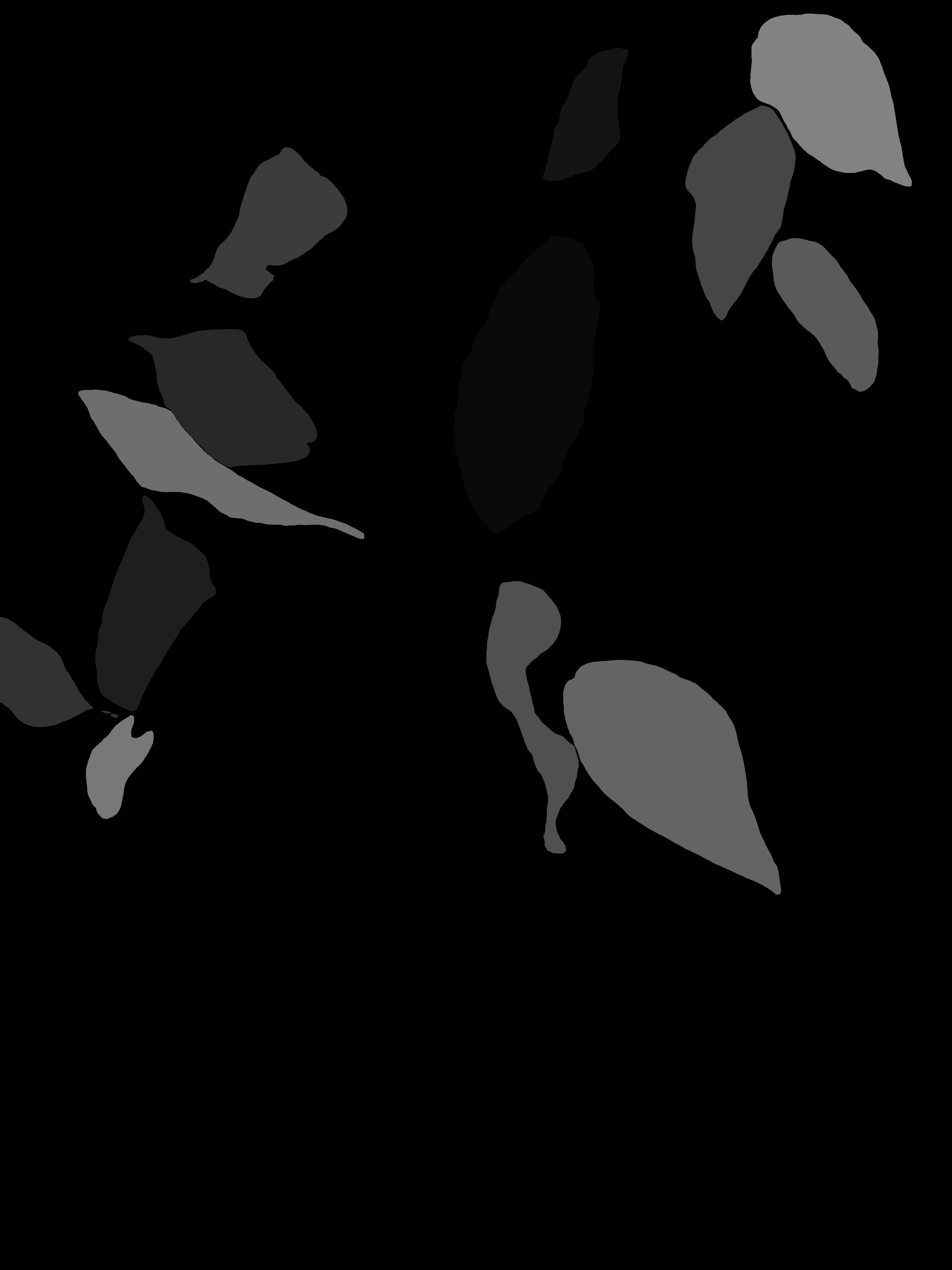}
    \captionof{figure}{Example leaf segmentation image generated by the SAM/ONNX Runtime}
\end{center}

\begin{center}
    \includegraphics[width=0.45\textwidth]{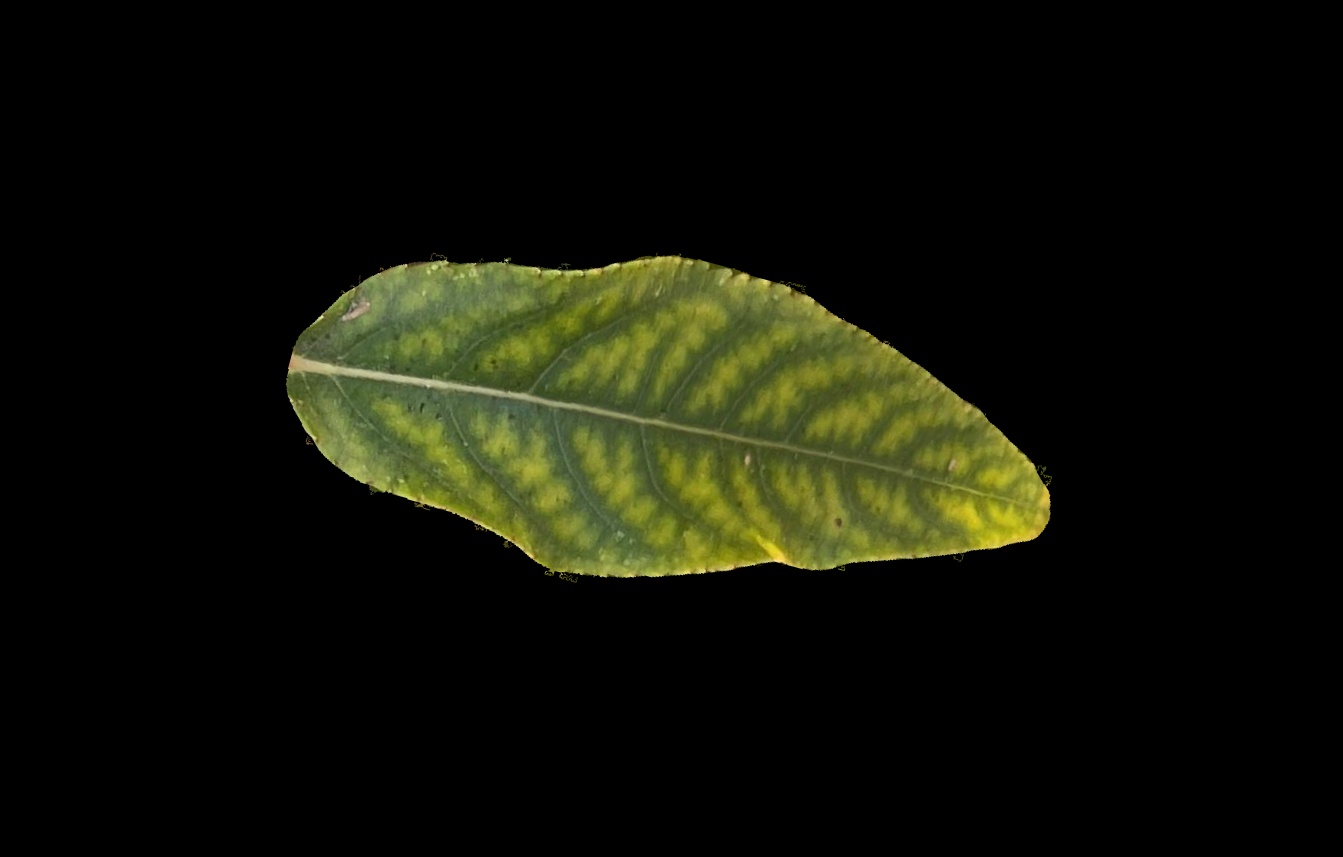}
    \captionof{figure}{Example leaf generated by cropping and rotating a filtered segmentation}
\end{center}

\section{Examples of Images with Obstructed Labels}
\label{app_blocked}

\begin{center}
    \includegraphics[width=0.40\textwidth]{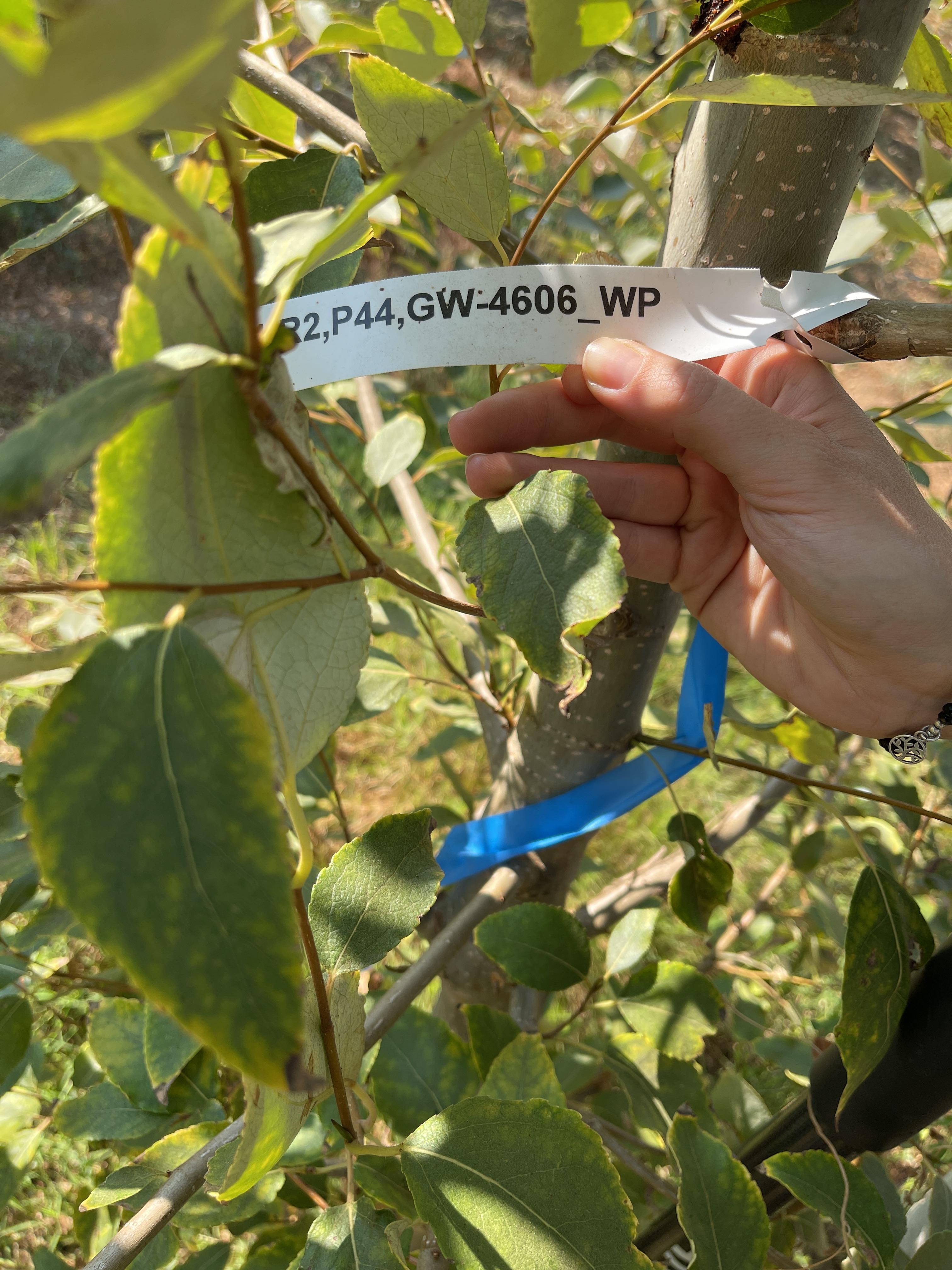}
    \captionof{figure}{Example of label blocked by a leaf}
\end{center}

\begin{center}
    \includegraphics[width=0.40\textwidth]{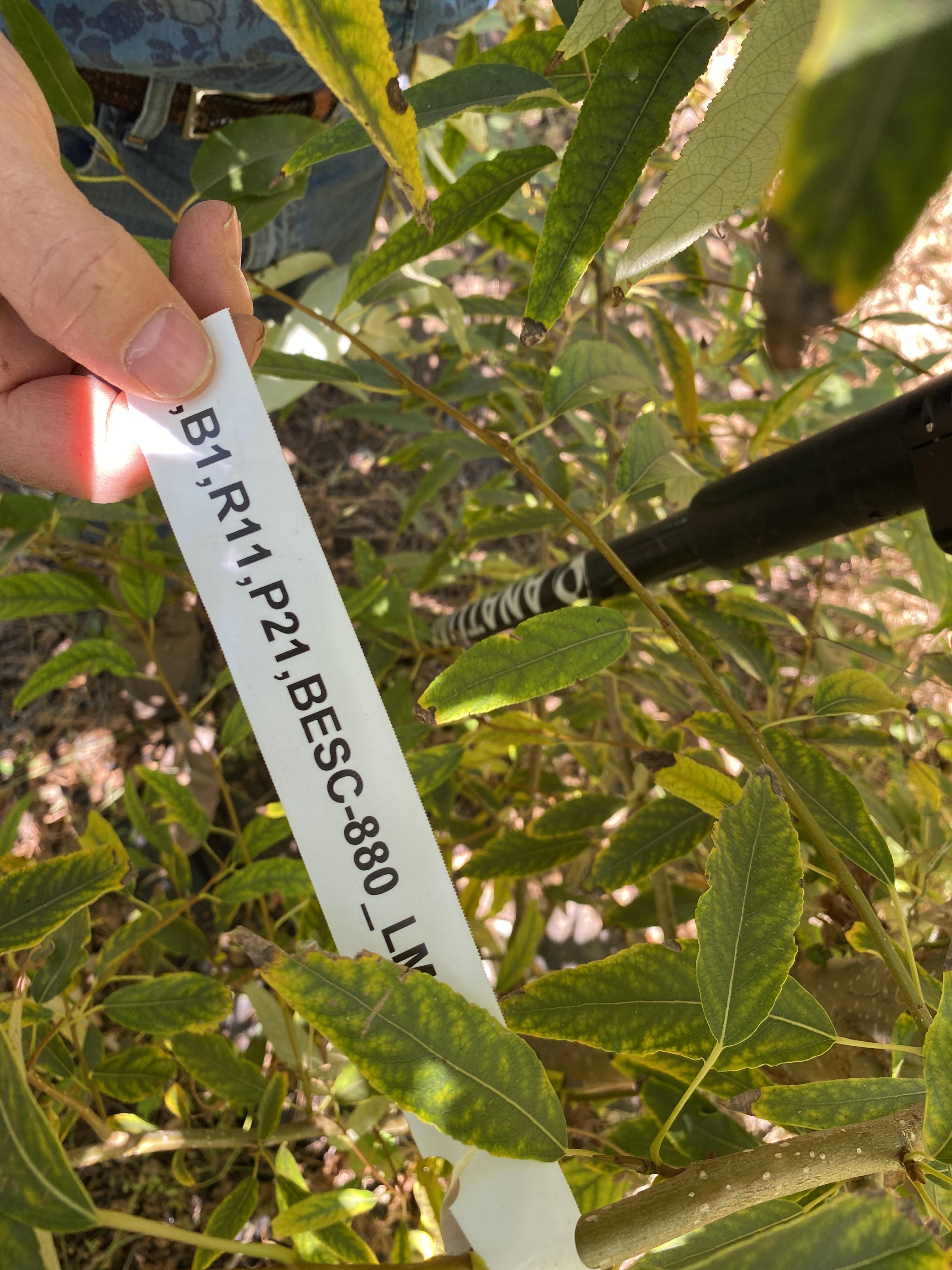}
    \captionof{figure}{Example of label blocked by the researcher's hand}
\end{center}

\begin{center}
    \includegraphics[width=0.40\textwidth]{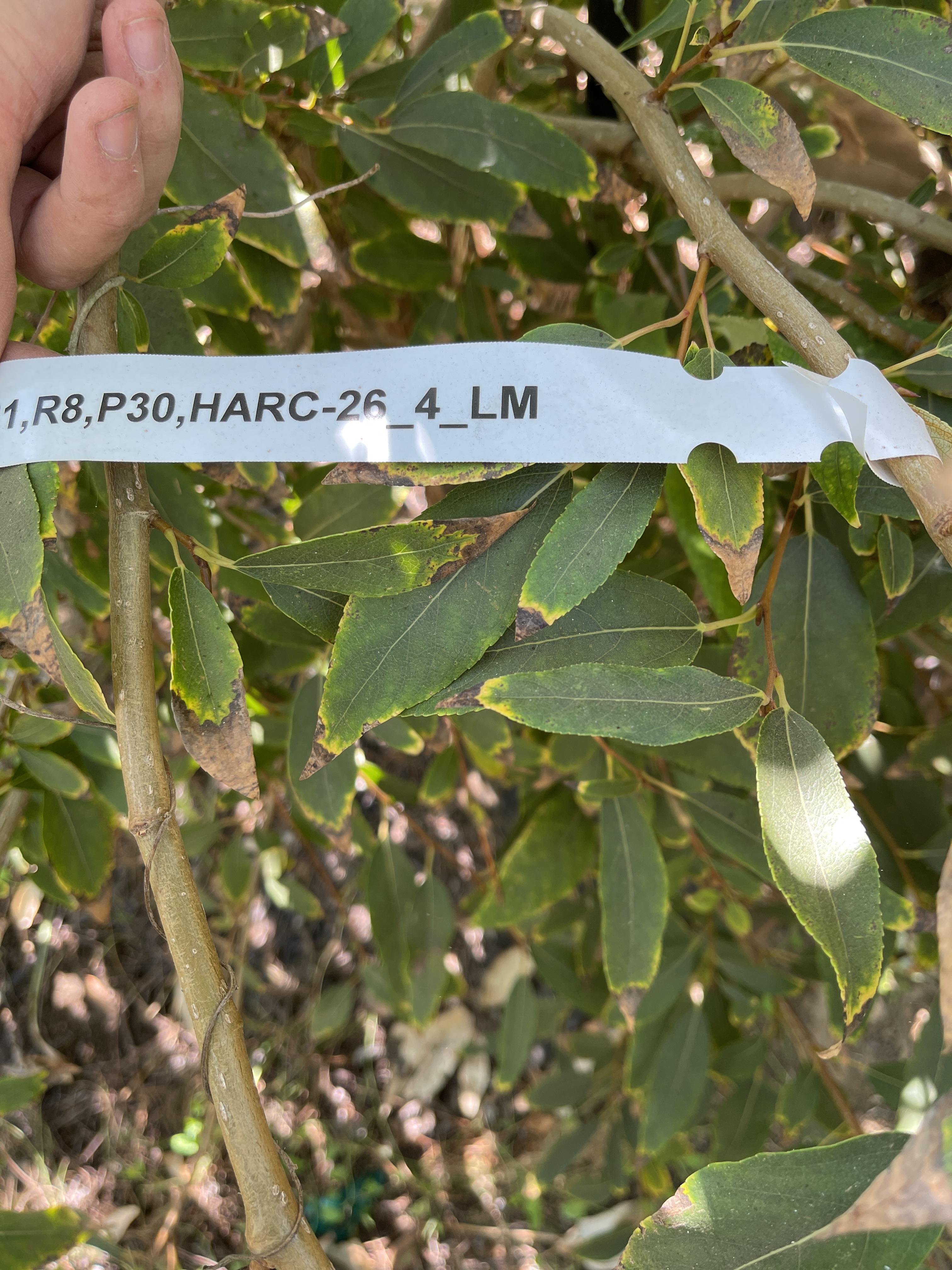}
    \captionof{figure}{Example of label cut by the boundary of the image}
\end{center}

\end{document}